# Set Interdependence Transformer: Set-to-Sequence Neural Networks for Permutation Learning and Structure Prediction

**Mateusz Jurewicz**[1,2] and **Leon Derczynski**[1]

[1] IT University of Copenhagen
[2] Tjek A/S
{maju, leod}@itu.dk

## Abstract

The task of learning to map an input *set* onto a permuted *sequence* of its elements is challenging for neural networks. Set-to-sequence problems occur in natural language processing, computer vision and structure prediction, where interactions between elements of large sets define the optimal output. Models must exhibit relational reasoning, handle varying cardinalities and manage combinatorial complexity. Previous attention-based methods require $n$ layers of their set transformations to explicitly represent $n$-th order relations. Our aim is to enhance their ability to efficiently model higher-order interactions through an additional interdependence component. We propose a novel neural set encoding method called the Set Interdependence Transformer[1], capable of relating the set's permutation invariant representation to its elements within sets of any cardinality. We combine it with a permutation learning module into a complete, 3-part set-to-sequence model and demonstrate its state-of-the-art performance on a number of tasks. These range from combinatorial optimization problems, through permutation learning challenges on both synthetic and established NLP datasets for sentence ordering, to a novel domain of product catalog structure prediction. Additionally, the network's ability to generalize to unseen sequence lengths is investigated and a comparative empirical analysis of the existing methods' ability to learn higher-order interactions is provided.

## 1 Introduction

There is a wide range of challenges where the objective is to find an optimal mapping from an unordered collection of objects to a permutation. This group of *set-to-sequence* tasks covers combinatorial optimization and structure prediction problems where exhaustive search is often not tractable, lending itself to neural network (NN) approaches.

Set-to-sequence challenges arise in many areas of application. Examples include natural language processing, in the form of sentence and paragraph ordering [Wang and Wan, 2019; Pandey and Chowdary, 2020], text comprehension [Li and Gao, 2020] and discourse coherence maximization [Farag, 2021]; computer vision for relative attribute learning [Santa Cruz and Fernando, 2017] and rigid point cloud registration [Yew and Lee, 2020]; reinforcement learning for managing the combinatorial action space of an agent [Vinyals *et al.*, 2019]. For an overview, see Jurewicz and Derczynski (2021b). We adapt our method to a novel application domain in the form of predicting the structure of digital catalogs.

---
[1]Paper accepted for publication in the IJCAI-22 proceedings

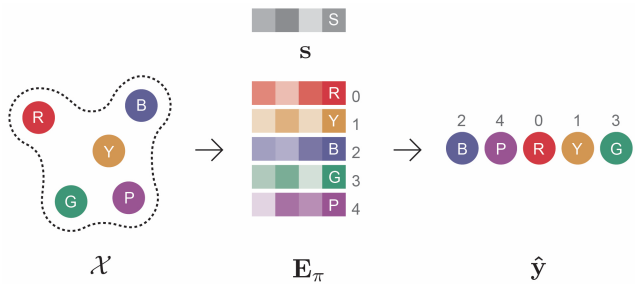

Figure 1: In a set-to-sequence task the initial set ($\mathcal{X}$) is passed to a set encoder, which obtains per-element representations ($\mathbf{E}_\pi$) and a permutation invariant representation ($\mathbf{s}$) of the entire set. A permutation decoder uses them to order the elements into a sequence ($\hat{\mathbf{y}}$).

Set-to-sequence models can be thought of as consisting of two distinct parts (see figure 1). Firstly, the set encoder obtains representations of both the elements of a set individually and of the set in its entirety. Secondly, a permutation learning module uses these two representations to predict a reordered sequence. Each stage presents unique challenges. Set-input methods are required to handle inputs of any dimensionality due to the examples being sets of varying cardinalities [Lee and Lee, 2019]. Additionally, the learned representation of a given set of cardinality $n$ must be identical for all $n!$ possible permutations of the vector that represents it. This permutation invariance can be achieved through a variety of symmetric functions, from simple summation [Zaheer *et al.*, 2017], through self-attention [Lee and Lee, 2019] to the output of a bipartite matching algorithm [Skianis and Konstantinos, 2020]. Solving challenging set-to-sequence problems requires a degree of relational reasoning, thus these functions benefit from being capable of encoding higher-order interac-tions between set elements [Huang *et al.*, 2020].

Predicting a permutation is also a challenge by itself. One of the main difficulties in dealing with combinatorial objects is that the number of possible output sequences increases factorially in the cardinality of the set. Additionally, the space of all possible permutations is not smooth, preventing direct use of gradient-based methods without a relaxation of the concept [Diallo *et al.*, 2020]. Finally, when the target is a complex structure represented by a permutation, it can be difficult to obtain evaluation and loss functions that are only sensitive to meaningful alterations of this structure.

To address these challenges, we propose a novel set encoding method which, unlike its predecessors, jointly learns the permutation invariant representation of the entire set and the permutation equivariant representations of individual set elements. Whilst methods that rely entirely on a pooled representation of the set have been applied to set-to-sequence problems [Vinyals *et al.*, 2016], their performance drops sharply as the cardinality of the input set increases, compared to methods that obtain both types of representations [Wang and Wan, 2019; Yin *et al.*, 2020]. To our knowledge, no other NN set encoding method learns the representations of set elements and the entire set jointly through an adjusted attention mechanism. Instead, they obtain the permutation equivariant representations of elements and then pool them through various symmetric operations to derive the encoding of the set proper. In this paper, we show empirically that learning them jointly is advantageous for the purposes of permutation learning and structure prediction.

We propose a complete, 3-part neural network architecture designed for performing the set-to-sequence mapping on inputs of any cardinality, consisting of an initial set encoder, an interdependence encoder and a permutation module. We then showcase its usefulness on a number of challenges, ranging from toy problems such as the Travelling Salesman Problem, through learning context-free and context-sensitive grammars to robust tasks such as sentence ordering and the novel task of catalog structure prediction. We also demonstrate our model's ability to generalize to unseen lengths and empirically prove its ability to learn higher-order interaction rules on a number of easily customized, synthetic structure datasets and via corresponding evaluation functions. All code, hyperparameters and datasets required for repeated experiments are provided in the supplement.

Our main contributions are summarized as follows:

- A novel, fully differentiable set encoding method designed specifically for permutation learning and structure prediction challenges, capable of learning higher-order interactions within sets of any cardinality in a single layer of the proposed transformations.
- A complete set-to-sequence model outperforming state-of-the-art methods on established datasets and within the novel application domain of catalog structure prediction.
- An expanded library for generating synthetic set-to-sequence structure datasets. In addition, we provide easy-to-use tools for obtaining detailed performance reports through customizable metrics, which allow researchers to measure and empirically confirm a model's relational reasoning capabilities.

## 2 Model

The goal is to transform an input set of any cardinality into the permuted sequence of its elements. To do this, the proposed set-to-sequence model consists of three core components: (i) a basic set encoder, (ii) a novel *interdependence encoder* and (iii) a permutation decoder. The initial encoder uses a learned pooling function to obtain (a) the permutation equivariant representations of individual set elements, and subsequently (b) the permutation invariant representation of the entire set. These two representations are then transformed together in the interdependence encoder, such that higher order interactions between both individual set elements and the set in its entirety can be learned in a single step. Finally, the permutation decoder sequentially selects the elements to form the output sequence by using these two representation via an enhanced pointer attention mechanism. An overview of our complete interdependence architecture is shown in Figure 2.

From a formal standpoint, the model is given a set $\mathcal{X}$ of any cardinality $n$, consisting of individual set elements represented by fixed-length vectors $\mathbf{x}_i$ of dimensionality $d$, in the form of a matrix $\mathbf{X}_\pi$ arbitrarily ordered according to some permutation $\pi$:

$$\mathcal{X} = \{\mathbf{x}_1, \ldots, \mathbf{x}_n\} \sim \mathbf{X}_\pi \in \mathbb{R}^{n \times d} \qquad (1)$$

The task is then to sequentially select individual set elements $\mathcal{X}$ in the target order, represented by a vector of indices $\mathbf{y} \in \mathbb{N}^{1 \times n}$ referencing the original placement of set elements in the $\mathbf{X}_\pi$ matrix. This process continues until there are no remaining candidate elements.

### 2.1 Basic Set Encoder

The initial encoder necessarily consists of learned function $f_e$ for transforming set elements in a permutation equivariant way, and learned function $f_s$ for pooling those element representations ($\mathbf{E}_\pi$) into a permutation invariant embedding of the entire set ($\mathbf{s}$), such that:

$$f_e(\mathbf{X}_\pi) = (\mathbf{e}_{\pi(1)}, \ldots, \mathbf{e}_{\pi(n)}) = \mathbf{E}_\pi \qquad (2)$$

$$\forall \pi \in \Pi \left( (f_s \circ f_e)(\mathbf{X}_\pi) = \mathbf{s} \right) \qquad (3)$$

In our proposed model this basic set encoder takes the form of a simplified Set Transformer [Lee and Lee, 2019], which we chose for its ability to explicitly represent inter-item interactions. We denote initial transformer-style attention toward permutation equivariant element representations $\mathbf{E}_\pi$ as:

$$\text{Att}(\mathbf{Q}, \mathbf{K}, \mathbf{V}) = \text{softmax}\left(\frac{\mathbf{Q}\mathbf{K}^\top}{\sqrt{d_k}}\right)\mathbf{V} \qquad (4)$$

where $\mathbf{Q}$, $\mathbf{K}$, $\mathbf{V}$ are the projected query, key and value matrices obtained from $\mathbf{X}_\pi$ via weight matrices $\mathbf{W}^Q$, $\mathbf{W}^K$ and $\mathbf{W}^V$; $\sqrt{d_k}$ is the standard transformer normalizing factor.

This operation is repeated in a multi-head fashion for each of the $m$ heads, whose outputs are concatenated and further transformed via learned weight matrix $\mathbf{W}^O$, without positional encoding or dropout:

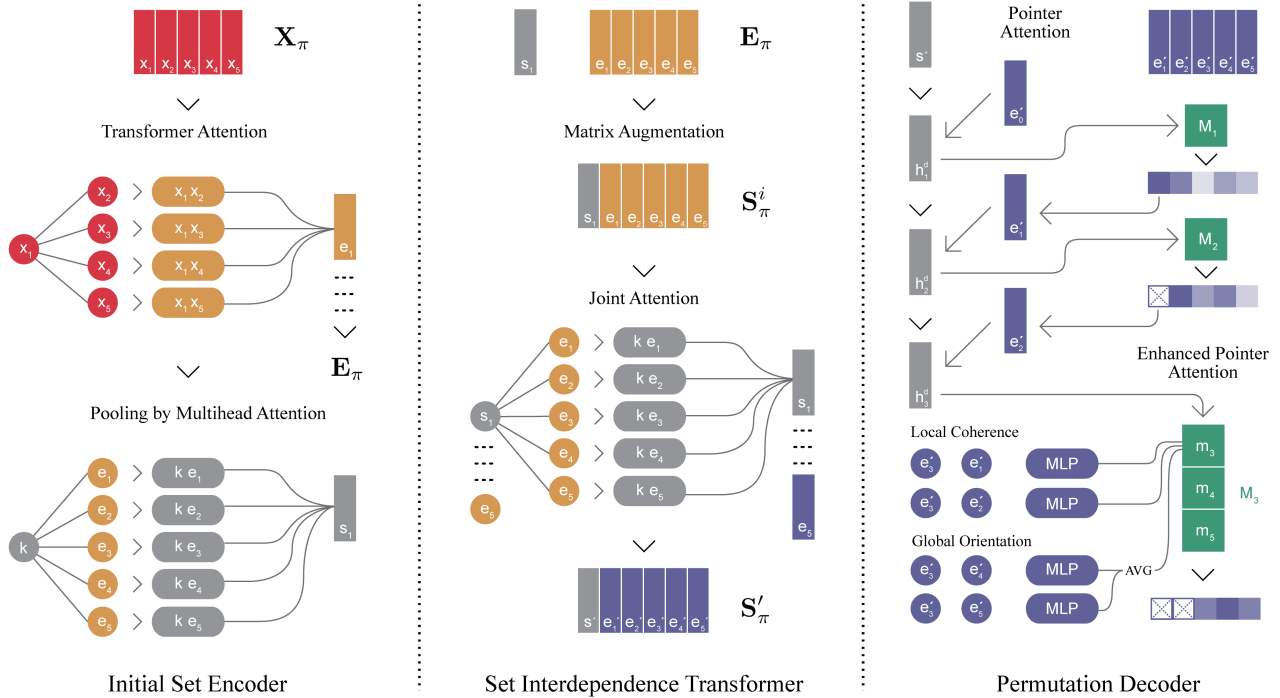

Figure 2: Three stages of the proposed set-to-sequence model. The Initial Set Encoder (left) obtains a permutation equivariant element representation ($\mathbf{E}_\pi$) through transformer-style attention. It then learns the permutation invariant encoding ($\mathbf{s}$) of the entire set via Pooling by Multihead Attention. The Set Interdependence Transformer (middle) augments the per-element matrix with the set encoding into $\mathbf{S}_\pi$ and performs further self-attention transformations, allowing for higher order interactions being modelled between set-elements and the set in its entirety. Finally, a Permutation Decoder (right) is used to sequentially select the next element in the output sequence until there are none left.

$$\mathbf{H}_i = \text{Att}(\mathbf{X}_\pi \mathbf{W}_i^Q, \mathbf{X}_\pi \mathbf{W}_i^K, \mathbf{X}_\pi \mathbf{W}_i^V) \quad (5)$$

$$\mathbf{E}_\pi = \text{Concat}(\mathbf{H}_1, \ldots, \mathbf{H}_m) \mathbf{W}^O \quad (6)$$

To obtain a permutation invariant representation ($\mathbf{s}$) of the set we apply simplified Pooling by Multihead Attention [Lee and Lee, 2019], which performs an attention transformation between a learned seed vector $\mathbf{k} \in \mathbb{R}^{1 \times d}$ as the query and $\mathbf{E}_\pi$ as the key and value for each of the $m$ heads:

$$\mathbf{s}_j = \text{Att}(\mathbf{k}_j \mathbf{W}_j^Q, \mathbf{E}_\pi \mathbf{W}_j^K, \mathbf{E}_\pi \mathbf{W}_j^V) \quad (7)$$

$$\text{PMA}(\mathbf{k}, \mathbf{E}_\pi) = \text{Concat}(\mathbf{s}_1, \ldots, \mathbf{s}_m) \mathbf{W}_s^O = \mathbf{s} \quad (8)$$

### 2.2 Set Interdependence Transformer

At this stage, we could rely on the transformations that lead to $\mathbf{E}_\pi$ to encode the dependencies between elements of $\mathcal{X}$. However, a single layer of such transformations can only explicitly capture *pairwise* relations, as it computes attention between pairs of elements [Lee and Lee, 2019]. Thus it would require up to $n$ stacks to explicitly encode higher-order interactions among an entire set of cardinality $n$. Our proposed set encoder allows efficient capture of dependencies between set elements and the set in its entirety through adjusted transformer-style attention. We refer to it as the Set Interdependence Transformer, or **SIT**.

The SIT performs attention-based transformations between both individual set elements and the permutation-invariant representation of the set itself in the form of an augmented matrix $\mathbf{S}_\pi$. The initial permutation invariant representation of the entire set $\mathbf{s}$ is treated as an element of a new set:

$$\mathbf{S}_\pi^i = (\mathbf{E}_\pi | \, \mathbf{s}) \quad (9)$$

$$\text{SIT}_i(\mathbf{s}, \mathbf{E}_\pi) = \sigma\left(\frac{(\mathbf{S}_\pi^i \mathbf{W}_i^Q)(\mathbf{S}_\pi^i \mathbf{W}_i^K)^\top}{\sqrt{d_\mathbf{s}}}\right) \mathbf{S}_\pi^i \mathbf{W}_i^V \quad (10)$$

$$\text{SIT}_i(\mathbf{s}, \mathbf{E}_\pi) = \mathbf{S}_\pi' = (\mathbf{E}_\pi' | \, \mathbf{s}') \quad (11)$$

where $d_\mathbf{s}$ is a scaling factor equal to the length of the permutation invariant vector representing the entire set, $\sigma$ is a non-linear activation function and $\mathbf{W}$ matrices are learned parameters. The above operation can be repeated over any number of heads as described under the inital basic encoder (subsection 2.1.

This method enables encoding of dependencies between individual set elements and the set in its entirety. Even a single SIT layer is capable of modelling higher-order interactions. This is beneficial for tasks over large sets, such as point cloud challenges, where the cardinality is prohibitively high for pairwise transformations.

To obtain the separate set and set-element representations required by a permutation module, we reverse the augmentation and retrieve the transformed $\mathbf{s}'$ vector by its index. Note that the required permutation invariance of $\mathbf{s}'$ is maintained during the permutation equivariant $\text{SIT}_i$ transformations.

### 2.3 Permutation Decoder

Finally, we use the $\mathbf{E}'_\pi$ and $\mathbf{s}'$ representations in a permutation decoder employing enhanced pointer-style attention [Yin *et al.*, 2020]. This takes the form of an LSTM-based pointer network with two additional mechanisms for pairwise ordering predictions towards improved global and local coherence of the output sequence. Formally, we calculate the conditional probability of a predicted order $\hat{\mathbf{y}}$ as:

$$p_\theta(\hat{\mathbf{y}} \mid \mathcal{X}) = \prod_{i=1}^{n} p_\theta(\hat{y}_i \mid \hat{\mathbf{y}}_{<i}, \mathbf{E}'_\pi, \mathbf{s}') \tag{12}$$

$$p_\theta(\hat{y}_i \mid \hat{\mathbf{y}}_{<i}, \mathcal{X}) = \text{softmax}(\mathbf{v}^\top \tanh(\mathbf{W}_1 \mathbf{h}_i^d + \mathbf{W}_2 \mathbf{M}_i)) \tag{13}$$

$$\mathbf{h}_i^d = \text{LSTM}(\mathbf{h}_{i-1}^d, \mathbf{e}'_{i-1}) \, , \, \mathbf{h}_0^d = \mathbf{s}' \tag{14}$$

where $\mathbf{v}$, $\mathbf{W}_1$ and $\mathbf{W}_2$ are model parameters, $n$ is the set's cardinality, tanh is the hyperbolic tangent nonlinearity, $\mathbf{e}'_{i-1}$ is the embedding of the set element selected at previous step $i-1$ and $\mathbf{h}_i^d$ is the hidden state of the permutation module at current step $i$. The first hidden state $\mathbf{h}_0^d$ is initialized from the permutation invariant set representation $\mathbf{s}'$. The $\mathbf{M}_i$ matrix provides additional context consisting of two types of information. The first is global orientation relating all remaining unordered set elements to each other and the second is local coherence between previously selected elements and remaining candidates. This context is obtained through "history" and "future" sub-modules from $\mathbf{E}'_\pi$. The sub-modules make pairwise ordering predictions in relation to each candidate element, which are then concatenated to form $\mathbf{M}_i$. For specific implementation details, see Yin *et al.* (2020).

During training, given a batch $\mathcal{B}$ of $n$ examples of the form $(\mathcal{X}_i, \mathbf{y}_i)$, we minimize the loss function:

$$\mathcal{L}(\theta) = -\frac{1}{n} \sum_{(\mathcal{X}, \mathbf{y}) \in \mathcal{B}} (\log p_\theta(\mathbf{y} \mid \mathcal{X}) + \lambda \, \mathcal{L}_s) \tag{15}$$

where $\theta$ is the set of all model parameters and $\lambda = 0.1$ is a hyperparameter that balances the first term of the loss with $\mathcal{L}_s$, a cross-entropy loss of the future and history sub-modules.

## 3 Experiments

All datasets, hyperparameters and code are freely available and described in detail in the linked supplementary material. The provided code includes any and all data pre-processing and generation steps, where applicable. Full requirements are explicitly listed. Best performance is reported in **bold**, second best is underlined. Reported results are averaged over 3 full training runs, standard deviation is reported after the ± sign.

### 3.1 Setup

We compare all models on the planar symmetric Travelling Salesman Problem (TSP) [Vinyals *et al.*, 2015] and 3 types of context-free and context-sensitive grammars, as well as the ROCStory sentence ordering dataset [Mostafazadeh and Chambers, 2016] and the PROCAT product catalog dataset [Jurewicz and Derczynski, 2021a], following provided train, test and validation splits. Additionally, we expand upon the synthetic structure dataset from PROCAT and report performance per $n$-th order rule of interaction required.

We use Kendall's Rank Correlation Coefficient ($\tau$) and Perfect Match Ratio (PMR) as primary metrics, scaled by a factor of a hundred for readability, following convention [Wang and Wan, 2019; Yin *et al.*, 2020; Pandey and Chowdary, 2020]. For TSP we report average tour length (shorter is better) and in both grammars and synthetic structure experiments the percentage of valid predictions per ruleset. For the representations of natural language entities in ROCStory and PROCAT we use the concatenated and averaged output of the last 4 layers of the cased large version of BERT [Devlin *et al.*, 2019], frozen during training to isolate the effect of different set encoding methods on the permutation task performance. The models' exact layer dimensions are given in the supplement, with the number of learnable parameters of each model varying by less than 5% per task. The proposed model consisted of 3-5 layers of SIT transformations over 4 attention heads, with the hidden size of 256. The AdamW [Loshchilov and Hutter, 2017] optimizer was used with weight decay coefficient 1e-2, learning rate ($\alpha$) 1e-4, dropout rate of 0.1 and batch size 32, for 50 epochs. All experiments were performed on cloud-based GPU instances, with NVIDIA Quadro P6000 graphics cards (24 GB) and 8 CPU cores.

### 3.2 Baselines

Our proposed model is compared with state-of-the-art set encoders combined with the same permutation component:

- **DeepSets**: *Zaheer et al.* [2017] encodes set elements independently and identically via a fully connected layer and then sums them.
- **Set Transformer**: *Lee et al.* [2019] uses transformer-style attention in a learned pooling by multihead attention function.
- **AttSets**: *Yang et al.* [2020] uses convolutional weighted attention and sum pooling.
- **RepSet**: *Skianis et al.* [2020] obtains permutation invariance through bipartite graph matching.

and with several complete set-to-sequence models:

- **Pointer Network**: *Vinyals et al.* [2015] consists of an RNN set encoder and basic pointer-style attention without pairwise predictions.
- **Read-Process-Write**: *Vinyals et al.* [2016] obtains permutation invariance via an adjusted LSTM network, which performs a number of passes over the input set.
- **ATTOrderNet**: *Vinyals et al.* [2018] combines transformer-style attention for set encoding with layer normalization, average pooling and a basic pointer-style permutation module.
- **Enhanced PtrNet**: *Yin et al.* [2020] introduces the global and local context sub-modules, adding pairwise judgments to the permutation mechanism.

|  | ROCStory | | PROCAT | | Synthetic $|y|=30$ | |
| --- | --- | --- | --- | --- | --- | --- |
| Method | PMR | Kendall's $\tau$ | PMR | Kendall's $\tau$ | PMR | Kendall's $\tau$ |
| DeepSets (2017) | 33.81 ± 4.55 | 62.41 ± 4.2 | 15.21 ± 2.69 | 35.15 ± 3.8 | 25.61 ± 1.90 | 40.94 ± 2.7 |
| Set Transformer (2019) | 41.94 ± 1.29 | 73.15 ± 1.9 | 21.03 ± 0.98 | 42.74 ± 2.6 | 30.23 ± 1.86 | 44.71 ± 2.9 |
| AttSets (2020) | 42.51 ± 1.45 | 74.81 ± 2.3 | 19.24 ± 1.30 | 38.44 ± 1.7 | 31.48 ± 2.04 | 46.20 ± 3.2 |
| RepSet (2020) | 42.47 ± 1.61 | 73.39 ± 1.7 | 22.72 ± 1.19 | 41.30 ± 2.5 | 34.84 ± 1.99 | 47.95 ± 3.4 |
| PointerNet (2015) | 28.73 ± 3.91 | 59.72 ± 6.2 | 02.90 ± 1.17 | 16.85 ± 3.4 | 17.33 ± 3.41 | 32.66 ± 3.1 |
| Read-Process-Write (2016) | 20.38 ± 7.22 | 51.78 ± 8.5 | 03.54 ± 1.06 | 21.11 ± 2.9 | 21.40 ± 2.07 | 36.28 ± 4.3 |
| ATTOrderNet (2018) | 41.14 ± 2.10 | 73.02 ± 2.0 | 17.33 ± 2.31 | 37.47 ± 3.2 | 28.02 ± 2.40 | 42.67 ± 3.4 |
| Enhanced PointerNet+ (2020) | 44.32 ± 1.25 | 76.43 ± 1.3 | 22.09 ± 1.57 | 42.53 ± 1.9 | 34.34 ± 1.39 | 48.14 ± 2.9 |
| Set Interdep. Transformer | **47.00 ± 0.89** | **79.86 ± 0.9** | **25.61 ± 1.81** | **46.41 ± 1.3** | **37.16 ± 1.01** | **52.03 ± 2.2** |

Table 1: Precise Match Ratio (PMR) and Kendall's Rank Correlation Coefficient ($\tau$) results for ROCStory, PROCAT and the Synthetic task.

|  | TSP cardinality | | |
| --- | --- | --- | --- |
| Method | $n=10$ | $n=15$* | $n=20$* |
| Held–Karp | 2.97 | 3.87 | 4.42 |
| Random | 4.48 | 7.32 | 8.96 |
| DeepSets | 3.27 ± 0.12 | 4.37 ± 0.19 | 5.12 ± 0.14 |
| SetTrans | 3.00 ± 0.02 | 3.99 ± 0.05 | 4.55 ± 0.08 |
| AttSets | 3.06 ± 0.06 | 4.02 ± 0.13 | 4.64 ± 0.19 |
| RepSet | 3.03 ± 0.05 | 4.13 ± 0.04 | 4.59 ± 0.11 |
| PtrNet | 3.18 ± 0.26 | 4.85 ± 0.69 | 5.83 ± 0.85 |
| RPW | 3.67 ± 0.10 | 4.92 ± 0.04 | 4.82 ± 0.11 |
| ATTOrdNet | 3.10 ± 0.06 | 4.53 ± 0.15 | 5.63 ± 0.39 |
| PtrNet+ | 3.01 ± 0.04 | 3.92 ± 0.06 | 4.52 ± 0.09 |
| SIT | **2.98 ± 0.01** | **3.90 ± 0.03** | **4.46 ± 0.06** |

Table 2: Performance on the TSP in terms of average tour length.

### 3.3 Core Results

As shown in Table 2, our proposed model outperforms baselines on the TSP by predicting shorter average paths and with smaller standard deviation. The improvement is more pronounced as input set cardinalities increasingly differ from those seen during training, unseen-cardinality sets being marked with an asterisk. Specifically, the proposed model predicts paths that are shorter than the second-best by 0.02 and 0.06 when $n=15$ and $n=20$ respectively, exhibiting increased ability to generalize to unseen path node counts, even for sets of twice the cardinality.

Overall results on ROCStory and PROCAT are shown in Table 1. Our proposed model, SIT, outperforms the state-of-the-art on both datasets. PMR scores are increased by 2.68 and 2.89, and $\tau$ by 3.43 and 3.67, over second-best performances on the ROCStory and PROCAT datasets respectively.

We also compare all methods on synthetically-generated structures of length 30, following the default $n$-th order rulesets. Compared to the best benchmark, our method offers an improvement of 2.32 (PMR) and 1.89 ($\tau$). Table 3 presents the average percentage of predicted sequences that adhered to the rules underlying two context sensitive and one context free grammar, in the form of the Dyck language [Yu et al., 2019] consisting of $n \in [2,100]$ pairs of {} and (). The simplest grammar ($a^n b^n c^n$, $n$ sampled uniformly from [1..100]) was able to be fully learned by most models using the enhanced pointer network as their permutation module, our proposed method among them. With regards to the more challenging $a^n b^k c^{nk}$ context-sensitive grammar ($n$ and $k$ sampled uniformly from [1..25]) our method outperformed the second best by 1.24 % and by 1.56 % in the case of the Dyck language. Performance in terms of learning higher-order interaction rulesets is discussed in Section 3.4.

The main improvement of the proposed model architecture stems from the ability to relate the input elements to the set in its entirety, supplementing its capacity to encode higher-order interactions in the set representation, from which the first hidden state of the permutation decoder is initialized. Compared with the baseline set encoders and full set-to-sequence models, the combination of the SIT set encoder and the enhanced pointer network consistently outperforms on a plethora of tasks where the target permutation depends on the input set's composition. Additionally, in order to explicitly confirm the proposed model's ability to learn higher-than-pairwise interactions between set elements in a single layer of SIT transformation, we designed a comparative experiment on the expanded synthetic structure prediction task. Results are presented in Table 4 and discussed in the next section.

### 3.4 Relational Reasoning Study

To empirically compare a model's capacity to learn $n$-th order interactions between set elements, we expanded the synthetic structure library provided by Jurewicz and Derczynski (2021a). This task requires ordering a set of tokens into a target structure, defined by customizable rulesets which depend on $n$-th order relations between input set elements. Each benchmark model consisted of 5 stacked layers of its respective attention transformations (except the Set Transformer, which was tested with both 4 and 5 layers). The hypothesis is that even with learning only pairwise interactions per-layer, benchmark models should be able to learn the highest-order ruleset (5th). By comparison, the tested versions of our proposed set-to-sequence model had 2, 3 and 4 attention-based encoding layers, requiring them to learn higher-order inter-actions in a single SIT layer. Further details are given in the supplementary material.

|  | Formal Grammars | | |
| --- | --- | --- | --- |
| Method | $a^n b^n c^n$ | $a^n b^k c^{nk}$ | Dyck |
| DeepSets | 97.17 ± 2.9 | 94.52 ± 2.4 | 79.84 ± 2.3 |
| SetTrans | **100.0 ± 0.0** | 96.62 ± 0.8 | 92.16 ± 1.5 |
| AttSets | 98.24 ± 1.3 | 96.12 ± 1.1 | 92.31 ± 1.2 |
| RepSet | **100.0 ± 0.0** | 97.64 ± 0.9 | 93.22 ± 1.4 |
| PtrNet | 79.31 ± 5.3 | 75.85 ± 4.7 | 58.85 ± 6.7 |
| RPW | 86.41 ± 1.5 | 81.74 ± 1.2 | 61.13 ± 5.3 |
| ATTOrdNet | 97.75 ± 1.1 | 94.17 ± 1.0 | 84.31 ± 3.8 |
| PtrNet+ | **100.0 ± 0.0** | 96.82 ± 0.7 | 93.51 ± 1.2 |
| SIT | **100.0 ± 0.0** | **98.88 ± 0.6** | **95.07 ± 1.0** |

Table 3: Results for 2 context-sensitive and 1 context-free grammar (the Dyck language). Scores are between 0-100, reflecting what proportion of predicted sequences was grammatical.

|  | $n$-th Order Relation Ruleset | | |
| --- | --- | --- | --- |
| Method | $n = 3$ | $n = 4$ | $n = 5$ |
| DeepSets | 53.37 ± 6.1 | 40.12 ± 4.0 | 21.13 ± 0.2 |
| SetTrans (4) | 92.41 ± 1.6 | 92.90 ± 1.4 | 91.74 ± 1.7 |
| SetTrans (5) | 94.66 ± 1.9 | 93.38 ± 1.3 | 92.93 ± 1.5 |
| AttSets | 93.84 ± 2.8 | 92.93 ± 1.1 | 86.44 ± 1.3 |
| RepSet | 91.29 ± 3.4 | 90.84 ± 2.4 | 89.73 ± 2.9 |
| PtrNet | 41.26 ± 5.3 | 31.96 ± 4.8 | 15.23 ± 4.2 |
| RPW | 45.11 ± 2.0 | 36.31 ± 1.6 | 16.12 ± 2.5 |
| ATTOrdNet | 82.31 ± 4.4 | 67.07 ± 2.0 | 0.12 ± 1.4 |
| PtrNet+ | 89.58 ± 3.9 | 87.22 ± 3.2 | 86.94 ± 2.8 |
| SIT (2 layers) | 93.83 ± 3.6 | 89.01 ± 2.6 | 86.79 ± 1.9 |
| SIT (3 layers) | **98.73 ± 0.8** | 93.44 ± 2.3 | 92.72 ± 1.8 |
| SIT (4 layers) | 98.48 ± 1.2 | **97.52 ± 0.9** | **96.10 ± 1.6** |

Table 4: Synthetic structure prediction scores per ruleset type, split by order of interaction required. On a scale of 0-100, reflecting the proportion of valid predicted structures with regards to each ruleset.

As shown in Table 4, the 3-layer version of the proposed model has the highest score on the 3rd-order ruleset by a margin of 4.07. However, even the 2-layer model outperforms all but two benchmark methods, each consisting of 5 stacked layers. On the 4th-order ruleset the best result is obtained through the proposed model's 4-layer version; second-best performance is attained by the 3-layer model, showing its ability to learn higher-than-pairwise interactions in a single SIT layer. Similarly, on the 5th-order ruleset ($n = 5$ in the table) the 4-layer model outperforms the best benchmark by 3.07 percentage points, with a 5-layer Set Transformer followed by the same permutation component as our proposed set-to-sequence model achieving second best performance. As an ablation study, we include results for a 4-layer Set Transformer, the only difference between it and our 4-layer set-to-sequence model being the SIT matrix augmentation. Its presence accounts for a 4.36 performance increase.

## 4 Related Work

The earliest neural set-to-sequence model was proposed by Hopfield and Tank (1985). It introduced a constrained version of the set-to-sequence problem, in which the input set must be of a fixed cardinality. However, sets of varying cardinality require different neural architectures, making it limited in application. Kernels between sets have also been proposed to allow kernel methods, such as Support Vector Machines, to tackle set-input problems [Lyu, 2005]. These require a two-step approach where representation learning is separate.

Set-input NN methods, such as DeepSets [Zaheer *et al.*, 2017], obtain a permutation equivariant representation of the input set by encoding each element in an identical and independent way and summing them into a permutation invariant representation. The main limitation is an inability to explicitly model higher order interactions, which are lost during pooling. This is addressed by the Set Transformer [Lee and Lee, 2019], using self-attention; the AttSets [Yang and Wang, 2020], though weighted pooling with learned, feature-specific weights; and RepSet [Skianis and Konstantinos, 2020], by solving a series of network flow problems.

Complete set-to-sequence NN models include the Pointer Network [Vinyals *et al.*, 2015], later improved by its authors to obtain a permutation invariant representation of the set [Vinyals *et al.*, 2016], ATTOrderNet [Cui *et al.*, 2018] and the Enhanced Pointer Network, which requires the model to continuously make pairwise order predictions, resulting in increased sequence coherence [Yin *et al.*, 2020]. There also exists a plethora of other NN order prediction methods, ranging from listwise ranking approaches [Ai *et al.*, 2018] in the field of information retrieval (which, however, presume an existence of a query for which the relevant order is predicted), to permutation matrix approaches [Zhang *et al.*, 2019], which do not handle sets of varying cardinalities. Finally, reinforcement learning has also been applied to combinatorial optimization problems in conjunction with pointer-based attention [Kool *et al.*, 2018]. However, many interesting tasks, such as sentence ordering and complex structure prediction, do not provide the kind of fine-grained signal ideally required by these methods.

## 5 Conclusion

In this paper, we propose a novel set encoding method for permutation learning and structure prediction. The Set Interdependence Transformer (SIT) is capable of effectively learning higher-order interactions between input set elements, which is crucial for tasks requiring a degree of relational reasoning over large sets of varying cardinality. Our method can take input sets of any size and generalizes well to unseen lengths of the target output sequence. It is easily modularized and can be combined with an attention-based permutation decoder to form a complete set-to-sequence model. Experiments show that this architecture achieves state-of-the-art results on combinatorial optimization challenges, on NLP tasks such as sentence ordering, and in the novel domain of complex structure prediction in the context of product catalogs.

## Acknowledgements

This work was partly supported by an Innovation Fund Denmark research grant (number 9065-00017B) and Tjek A/S.